\documentclass[11pt,a4paper]{article}
\usepackage[hyperref]{emnlp2020}
\usepackage{times}
\usepackage{latexsym}

\usepackage{graphicx} 
\usepackage{pgfplots} 
\usepackage{amsmath} 
\usepackage{amssymb} 
\usepackage{hyperref}
\pgfplotsset{compat=1.16} 
\usepackage{subcaption} 
\usepackage{booktabs} 
\usepackage{multirow} 
\usepackage{fontawesome} 

\usepackage{microtype}

\aclfinalcopy 

\title{dictNN:\\ A Dictionary-Enhanced CNN Approach \\ for Classifying Hate Speech on Twitter}

\author{Maximilian Kupi, Michael Bodnar, \textbf{Nikolas Schmidt,}\and \textbf{Carlos Eduardo Posada} \\
Hertie School, Friedrichstraße 180, 10117 Berlin, Germany\\
\texttt{Contact: \href{mailto:m.kupi@phd.hertie-school.org}{m.kupi@phd.hertie-school.org}}}

\date{\today}

\begin{document}
\maketitle

\begin{abstract}
Hate speech on social media is a growing concern, and automated methods have so far been sub-par at reliably detecting it. A major challenge lies in the potentially evasive nature of hate speech due to the ambiguity and fast evolution of natural language. To tackle this, we introduce a vectorisation based on a crowd-sourced and continuously updated dictionary of hate words and propose fusing this approach with standard word embedding in order to improve the classification performance of a CNN model. To train and test our model we use a merge of two established datasets (110,748 tweets in total). By adding the dictionary-enhanced input, we are able to increase the CNN model's predictive power and increase the F1 macro score by seven percentage points.
\end{abstract}

\section{Introduction \& Related Work}\label{sec:introduction}
With the widespread use of social media, hateful comments – targeting individuals or groups based on ethnicity, national background, gender identity, sexual orientation, class or disability \cite[p.~130 ff.]{Jacobs2001HatePolitics} – have become a concerning issue \cite{Costello2018PredictorsYouth}. However, detecting and removing hate speech automatically is a complex task, due to the prevailing lack of a common hate speech definition, the nuances of natural language (e.g. sarcasm and double meaning), and the fast lexical evolution of hate speech \cite{MacAvaney2019}. In this respect, research has stressed the importance to distinguish between hateful and simply abusive or offensive speech in order to give both a differentiated orientation for manual tagging as well as a more nuanced basis for automated classification \citep{davidson2017,Fortuna2019ADataset, founta2018large}.

Although a variety of non-deep learning algorithms have been employed in past literature (see for example \citealp{Nobata2016AbusiveContent}; \citealp{Ibrohim2019Multi-labelTwitter}), neural networks have become the state-of-the-art for text classification and hate speech detection in recent years – in particular in combination with the use of various word-embedding techniques to represent the text data in a vector space \cite{kim_convolutional_2014, Kshirsagar2018PredictiveTwitter}.

Most current approaches include user data and other metadata in the analysis in order to improve the models' accuracy \cite{Mathew2019SpreadMedia, Ribeiro2018CharacterizingTwitter, Waseem2016HatefulTwitter, Miro-Llinares2018HateMicroenvironments, Stoop2019DetectingDevelop}. However, the focus of these models lies on identifying hateful user accounts and environments, rather than hateful content per se. Furthermore, the reliance on metadata impedes a model's applicability to other datasets or online platforms \cite{Meyer2019ADetector}.

Alternatively, scholars have explored the use of further lexical characteristics of hate speech, e.g. by making use of sentiment analysis \cite{Rodriguez2019AutomaticAnalysis, Schmidt2017, Watanabe2018HateDetection}.
Here, researchers have mostly relied on dictionary-based approaches \cite{Gitari2015ADetection, Mulki2019L-HSAB:Language, Martins2018, davidson2017}. Yet, to our best knowledge, these approaches have so far only been employed for machine learning that does not involve neural networks. 

We propose the usage of a dictionary approach in the preprocessing stage for a deep learning-based hate speech classification with CNNs. Our research contributions are the design of an additional dictionary-based tweet vectorisation approach and a method to combine the resulting vectors with standard word-embedding vectors. We also compare the performance of a simple CNN model architecture with and without applying our suggested approach on established hate speech datasets. The results show that our dictNN approach increases the model's performance.

\section{Proposed Method\protect\footnote{All code is accessible on \href{https://github.com/MaximilianKupi/nlp-project.git}{github.com}.}}\label{sec:proposed-methods}
\paragraph{Choice of model:}
CNN architectures have shown promising results in natural language processing \cite{Khan2020ANetworks}. They offer distinctive advantages, in particular for detecting hate speech, as they are adept at recognising patterns of word combinations \cite{Yin2016}. This specificity of CNNs is especially relevant in short texts like tweets, where certain combinations of words can be highly indicative for hate speech. Furthermore, CNN architectures have proven to be more resource efficient than their peers \cite{Adel2017}. For these reasons, we have chosen a CNN model architecture.

\paragraph{Enhancing the word embedding with input from a dictionary:}
As the key feature of our approach we enhance the input to the CNN with data derived from the \href{https://hatebase.org/}{\textit{Hatebase}} dictionary, an established crowd-sourced online dictionary of hateful terms. This dictionary contains over 3,500 hate terms in over 97 languages, and includes further information such as a term's average offensiveness score and whether or not a term is ambiguous. For our purposes we downloaded all English hate terms (1,530 in total) using the provided API. 

The dictionary-based input is meant to provide the model with additional information in cases where the tweets contain hateful terms. To calculate the vectors, each word in a tweet is looked up in the dictionary. In case a word occurs in the dictionary, its average offensiveness score – a number between 0 and 100 – is inserted at the respective vector location of the word. For unambiguously hateful terms the number of the score is doubled, since these terms most likely are particularly indicative of hate speech. All non-matching words are represented as zeros in the vector. In order to also match misspelled or \mbox{(self-)}censored hate words where certain letters have been replaced with symbols, we employ a difflib-based sequence comparison as suggested by \citet{Chiu2018}. Through experimental testing we determined the best similarity cut-off to be 0.85. In case a word matches multiple hate dictionary entries, the average of their offensive scores is used.

The derived vectors are then integrated into a vectorisation process of the tweets, which is implemented with a pretrained model of BERT \cite{Devlin2018}. BERT models are bidirectionally trained, so they achieve a better sense of language context than single-directional models. For reasons of efficiency and reliability we employ the BERT-base-uncased model \cite{Rajapakse2020}. Both vectorisation methods – our dictionary-based and the BERT vectorisation – are applied to every tweet in our dataset and the resulting vectors are stacked. Due to their differing lengths, the vectors of the offensiveness scores are stretched to match the length of the BERT vectors. The resulting matrix is then used as an input for the 2D-architecture of our CNN model, as opposed to the 1D-baseline-architecture which only uses BERT-vectorised word embeddings as input.

\begin{figure}[ht]
    \begin{center}
        \includegraphics[scale=0.8]{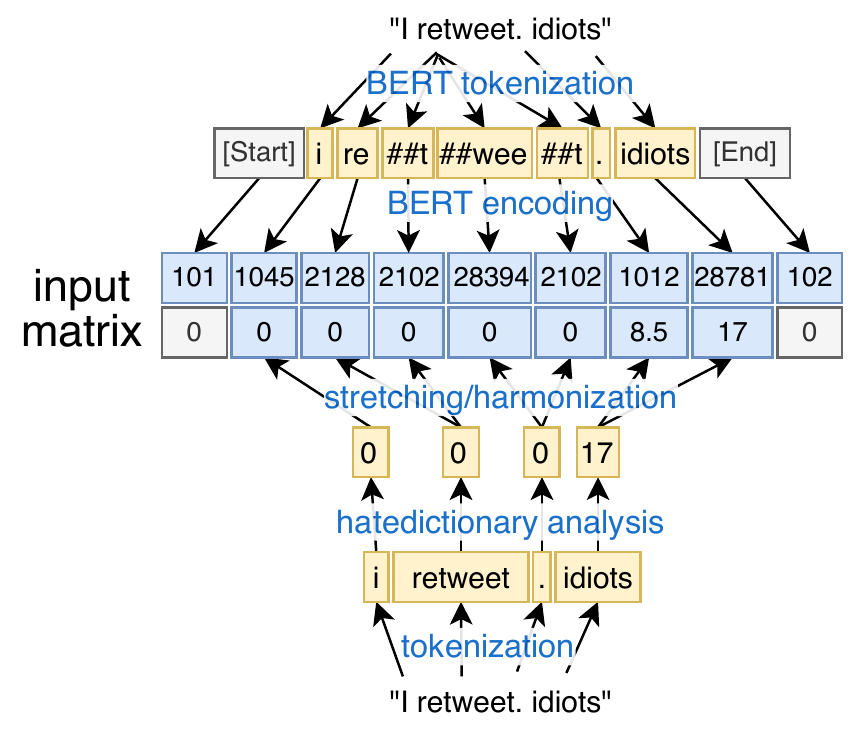}
        \caption{Preprocessing of tweets and combination of BERT vectorisation with hate dictionary approach.}
        \label{fig:vectorization}
    \end{center}
\end{figure}

\paragraph{Harmonising different vector lengths:}
While the BERT vectoriser produces a vector length that depends on the BERT specific tokenisation, our dictionary approach uses simple tokenisation without lemmatisation and other changes to best match the terms in the hate dictionary (see Figure~ \ref{fig:vectorization}).\footnote{The \href{https://hatebase.org/}{\textit{Hatebase}} dictionary contains only nouns. We tested different versions of our dictionary vectorisation – with and without lemmatisation – and found that lemmatising did not lead to better matching.} This produces two vectors, which can be of different lengths. To ensure that the value resulting from the hate dictionary analysis is located in the vicinity of the actual BERT-encoded word, we stretch the dictionary-derived vector with linear interpolation. This way, the length of both vectors match, and a word's hate dictionary evaluation will be close to its BERT embedding, thus adding additional information to the respective term in the second input dimension. 

To harmonise the input matrix length for our CNN over all tweets, the maximum vector length is set to 120 and padded with zeros.

\section{Experiments}\label{sec:experiments}
\paragraph{Model architecture:}
We use a CNN architecture with four hidden layers – three convolutional and one final linear layer – as well as a ReLU activation function, and batch normalisation after each convolution. The initial number of channels is set to one, as we input one matrix per tweet. The output channels of the first, second, and third CNN layer have been set to 16, 32, and 64 respectively. We apply a kernel of size three with zero padding and stride of one to preserve the feature dimensions throughout the network \cite{Dumoulin2016AGT}. The output features of the final linear layer have been set to three in order to match the three classes of the tweet annotation.\footnote{For a visualisation of the model architecture as well as a summary of the model parameters see Figure~\ref{fig: architecture} as well as Table~\ref{tab:model-1d} and \ref{tab:model-2d} in the \nameref{Appendix}.} 

\paragraph{Datasets:} We use the following two established datasets for our classification task: the dataset from \citet{davidson2017}, which is composed of 25,000 tweets that have been preselected using the crowd-sourced \textit{Hatebase} hate speech keyword lexicon (see Section~\ref{sec:proposed-methods}) and manually labeled using the crowd-sourcing platform CrowdFlower (labels: \textit{hate\_speech}, \textit{offensive\_language}, \textit{neither}); as well as the dataset from \citet{founta2018large}, which comprises 100,000 manually annotated tweets (again using CrowdFlower) whose annotation (labels: \textit{hateful}, \textit{abusive}, \textit{normal}, \textit{spam}) has been checked for statistical robustness. Since our focus does not lie on spam detection, we deleted all tweets in this category from the second dataset. Next, following the definitions of \citet{founta2018large}, according to which offensive and abusive language are two very similar categories, we merge both datasets in order to obtain a more balanced dataset as well as increase the sample size for the \textit{hate} class. Due to the still prevailing – yet improved, as compared to the original datasets – imbalance of the resulting dataset (110,748 tweets; see Figure~\ref{fig:counts} in the \nameref{Appendix}), we divided the data into training (70\%), validation (15\%), and test set (15\%), using a stratified shuffle split \citep{Scikit-learndevelopers2019a}.

\paragraph{Data preprocessing:} In order to remove noise before the vectorisation processes, the text was transformed to lower case, and symbols, spaces as well as standalone digits were removed. Tweet-specific traits, i.e. URLs, mentions, reserved words, emojis and smileys were also removed. The content of hashtags (i.e. "idiots" in the case of "\#idiots") as well as standard sentence punctuation (i.e. ?!.,) was preserved.

\paragraph{Soft- and hardware setup:} The project was coded and deployed in the programming language Python (Version 3.7), using the \citet{PyTorch2019} library. To train our models we used Nvidia Tesla K80 and T4 GPUs on  \href{https://colab.research.google.com/}{Google Colab}.

\paragraph{Evaluation method:} The loss was computed using a multi-class cross-entropy loss function. Apart from the usual loss and accuracy plots to ensure the correct training and a generalisability of the model, we evaluated the models based on the F1 macro score to account for the imbalanced class distribution in the dataset.\footnote{To calculate the evaluation metrics we use a \citet{Scikit-learndevelopers2019b} package.}

\paragraph{Experimental details:} To determine the best hyperparameters for our CNN model architecture, we made use of a grid search on the following discrete parameters: optimizer type (Adam, RMSprop, and SGD); learning rate (0.0001, 0.001, 0.01); whether to use a sampler in the data loader or class weights in the loss function to balance the dataset during training; and whether or not to use a scheduler to decrease the learning rate (factor 0.1) based on the performance of the F1 macro score. In case the class weights were used, the shuffle mode in the data loader was automatically set to true in order to ensure more robust training. During grid search we trained each model in the simple 1D version (without our dictionary-based preprocessing approach) for 45 epochs with a batch size of 16. The combinations of above-mentioned hyperparameters resulted in a total of 36 search attempts with an average runtime of 33 minutes. We then considered the results from the best epoch with respect to the F1 macro score.\footnote{The expected validation performance plot (Figure~\ref{fig:Expected Validation Performance}) can be found in the \nameref{Appendix}.} Based on the results from the grid search, the best hyperparameters were selected and the simple 1D version as well as the 2D version with additional dictNN preprocessing was trained for 90 epochs respectively. For those four runs the batch size was again set to 16 and the average run time was 67 minutes.\footnote{The F1 macro score as well as the loss and accuracy plots (Figures~\ref{fig:f1}, \ref{fig:loss}, \& \ref{fig:accuracy}) can be found in the \nameref{Appendix}.}

\paragraph{Results:} 
The best model from the grid search has the following specifications: Adam optimizer, 0.01 learning rate, dataset-balancing class weights in the loss function, and a deactivated scheduler. After retraining the 1D and 2D model for 90 epochs, the highest F1 macro score for the 1D model was yielded in epoch 29 and for the dictNN 2D model in epoch 36 respectively. Testing results of these models are shown in Table~\ref{tab: classification report}.\footnote{The classification report for each model's best epoch on the validation set (Table~\ref{tab: classification report on validation}) can be found in the \nameref{Appendix}.} As can be seen, our dictNN approach outperforms the simple 1D model on all levels. 

\noindent
\begin{table}[ht]
    \resizebox{\columnwidth}{!}{%
        \small\addtolength{\tabcolsep}{-3pt}
        \begin{tabular}{c r c c c c c}
        \toprule
        
        & \multicolumn{2}{c}{\bfseries Precision \hspace{0.05cm}} &
        \multicolumn{2}{c}{\bfseries Recall \hspace{0.05cm}} &
        \multicolumn{2}{c}{\bfseries F1 Score} \\
        \cmidrule(lr){2-3} \cmidrule(lr){4-5} \cmidrule(lr){6-7}
        & 1D & 2D \hspace{0.05cm} & 1D & 2D \hspace{0.05cm} & 1D & 2D \\
        \cmidrule(lr){1-7}
        \textit{hateful}    &  0.21 &  0.31 \hspace{0.05cm} &  0.26 &  0.30 \hspace{0.05cm} &  0.23 &  0.30 \\
        \textit{abusive}    &  0.69 &  0.76 \hspace{0.05cm} &  0.62 &  0.68 \hspace{0.05cm} &  0.66 &  0.72 \\
        \textit{normal}     &  0.73 &  0.77 \hspace{0.05cm} &  0.76 &  0.83 \hspace{0.05cm} &  0.74 &  0.80 \\
        \textbf{Accuracy}   &       &                       &       &                       &  0.68 &  0.74 \\
        \textbf{Macro Avg}  &  0.54 &  0.61 \hspace{0.05cm} &  0.55 &  0.61 \hspace{0.05cm} &  0.54 &  0.61 \\
        \textbf{Micro Avg}  &  0.68 &  0.74 \hspace{0.05cm} &  0.68 &  0.74 \hspace{0.05cm} &  0.68 &  0.74 \\
        \bottomrule
        
        \end{tabular}
    }
    \caption{Classification report of the best 1D model with BERT preprocessing and 2D model with additional dictNN preprocessing on the test set.}
    \label{tab: classification report}
\end{table}

\section{Analysis \& Ethical Considerations}\label{sec:analysis}
\begin{figure}[ht]
    \centering
    \begin{subfigure}{1\linewidth}
        \includegraphics[width=0.9\columnwidth]{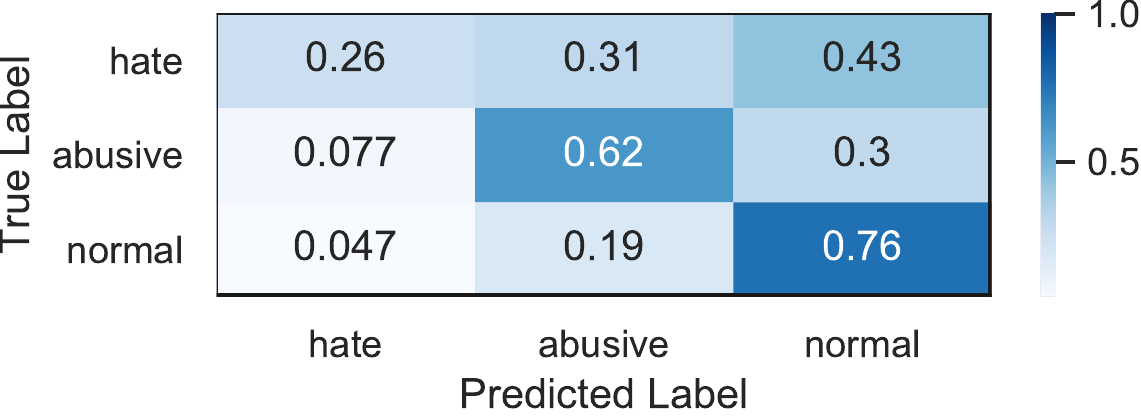} 
        \caption{1D Model with BERT preprocessing.}
        \label{fig:confusion:1d-bert}
    \end{subfigure}
        \begin{subfigure}{1\linewidth}
        \includegraphics[width=0.9\columnwidth]{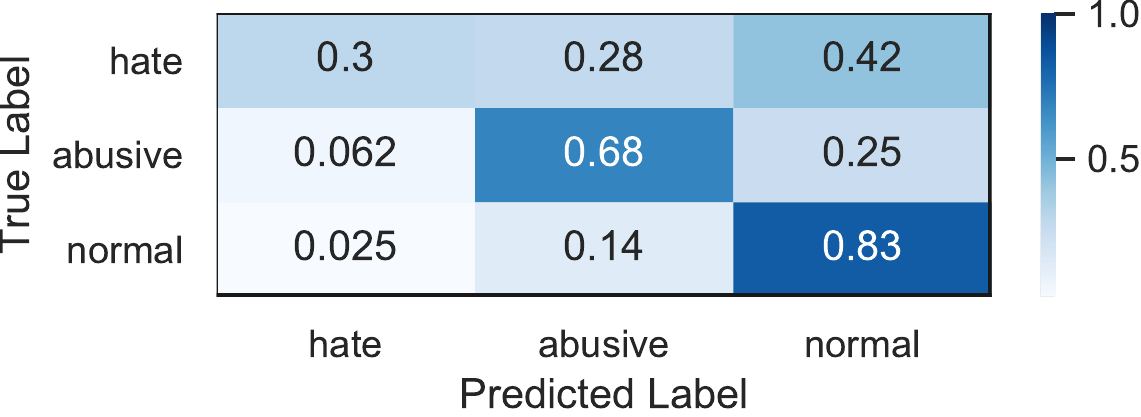}
        \caption{2D Model with additional dictNN preprocessing.}
        \label{fig:confusion:2d-dictnn}
    \end{subfigure}
    \caption{Confusion matrix of the 1D and the 2D model on the test set.}
    \label{fig:confusion}
\end{figure}

Comparing the two models' performances on the test set based on the confusion matrix (see Figure~\ref{fig:confusion}), shows the superiority of our approach over a simple BERT model.\footnote{The confusion matrix for each model's best epoch on the validation set (Figure~\ref{fig:confusion-on-validation-set}) can be found in the \nameref{Appendix}.} However, although the average hate scores per test set tweet are distributed as expected (i.e. \textit{hateful} 126.28, \textit{abusive} 89.91, and \textit{normal} 56.26), the still high percentage in the top right of the confusion matrix suggests that our approach does not substantially improve the misclassification of \textit{hateful} speech as \textit{normal} speech. 

Apart from the particularly skewed balance between these two classes, which leads to an over-representation and thus over-specification of the \textit{normal} tweets, a few other potential explanations account for this phenomenon: First, some tweets are labelled as \textit{hateful} but contain no hate words, so our dictionary-based vectorisation did not provide any additional information for the network to pick up (see first synthetic example test set tweet in Table~\ref{tab: Example tweets} of the \nameref{Appendix}). Second, some normal tweets have actually been wrongly labelled as \textit{hateful}, as they, for example, only report about hate (see example tweet 2). Third, our dictionary-preprocessing algorithm does not catch all the \textit{hateful} words in some tweets. This might be due to a not (yet) implemented \textit{bi}-gram matching or the term simply not occurring in the \textit{Hatebase} dictionary – as the dictionary only includes nouns. Example tweet 3 in Table~\ref{tab: Example tweets} fulfils both these conditions. 

Finally, the definition of hate speech and which tweets fall under this definition is still an ongoing debate \cite{Dorris2020TowardsLanguage}. Hence, training datasets might carry potential biases – particularly in connection with racial differences in language usage \citep{davidson-etal-2019-racial}. Although our dictionary-based preprocessing approach accounts for the potential ambiguity of hate terms by giving more weight to unambiguously hateful terms (see Section~\ref{sec:proposed-methods}), these possible biases are still an ethical caveat that has to be taken into account when applying our approach in the field.

\section{Conclusion}\label{sec:conclusion}
We have shown that enriching standard word embeddings with a dictionary-based embedding for hateful terms can substantially improve a CNN model's performance in distinguishing between \textit{hateful}, \textit{abusive}, and \textit{normal} tweets. Future research might further improve our approach, apply it to other classification problems or compare it to results of other state-of-the-art approaches.

\section{Acknowledgements}
The authors would like to express their gratitude to Antigoni M. Founta for granting access to the complete dataset of their study \cite{founta2018large}.

{\small
\bibliographystyle{acl_natbib}
\bibliography{emnlp2020.bib}
}

\newpage
~
\newpage 
\section{Appendix}\label{Appendix}

\begin{figure}[h]
    \begin{center}
        \includegraphics[width=0.45\textwidth]{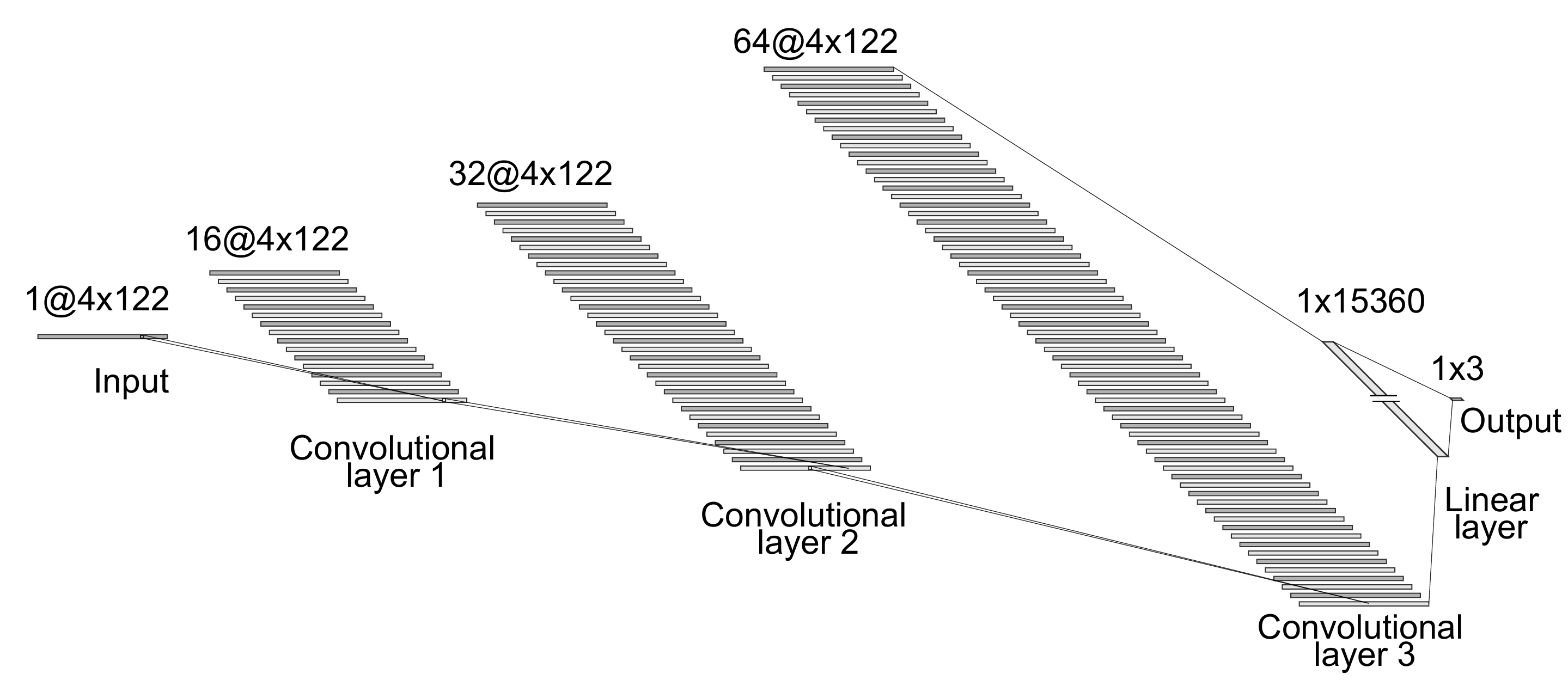}
        \caption{The employed CNN model architecture (in 1D and including padding).}
        \label{fig: architecture}
    \end{center}
\end{figure}

\begin{table}[h]
    \begin{tabular}{ l c r r }
        \toprule
        \addlinespace
        
        \bfseries Layer (type)    &  \bfseries    Output Shape    &  \bfseries Tr. Param \#    \\
        \hline
        Conv1d-1        &   [1, 16, 120]    &          64      \\
        BatchNorm1d-2   &   [1, 16, 120]    &          32      \\
        ReLU-3          &   [1, 16, 120]    &           0      \\
        Conv1d-4        &   [1, 32, 120]    &       1,568      \\
        BatchNorm1d-5   &   [1, 32, 120]    &          64      \\
        ReLU-6          &   [1, 32, 120]    &           0      \\
        Conv1d-7        &   [1, 64, 120]    &       6,208      \\
        BatchNorm1d-8   &   [1, 64, 120]    &         128      \\
        ReLU-9          &   [1, 64, 120]    &           0      \\
        Linear-10       &   [1, 3]          &      23,043      \\
        \bottomrule
        \addlinespace
    \end{tabular}
    \begin{center}
        Trainable params: 31,107 \\
        Non-trainable params: 0
    \end{center}
    
    \caption{1D model parameters.}
    \label{tab:model-1d}
\end{table}

\begin{table}[h!]
    \begin{tabular}{ l c r r }
        \toprule
        \addlinespace
        \bfseries Layer (type)    &  \bfseries    Output Shape     &   \bfseries Tr. Param \#    \\
        \hline
        Conv2d-1        &   [16, 16, 2, 120]    &         160       \\
        BatchNorm2d-2   &   [16, 16, 2, 120]    &          32       \\
        ReLU-3          &   [16, 16, 2, 120]    &           0       \\
        Conv2d-4        &   [16, 32, 2, 120]    &       4,640       \\
        BatchNorm2d-5   &   [16, 32, 2, 120]    &          64       \\
        ReLU-6          &   [16, 32, 2, 120]    &           0       \\
        Conv2d-7        &   [16, 64, 2, 120]    &      18,496       \\
        BatchNorm2d-8   &   [16, 64, 2, 120]    &         128       \\
        ReLU-9          &   [16, 64, 2, 120]    &           0       \\
        Linear-10       &            [16, 3]    &      46,083       \\
        \bottomrule
        \addlinespace
    \end{tabular}
    \begin{center}
        Trainable params: 69,603 \\ 
        Non-trainable params: 0
    \end{center}
    \caption{2D model parameters.}
    \label{tab:model-2d}
\end{table}

\begin{figure}[ht]
    \begin{center}
        \includegraphics[width=\columnwidth,keepaspectratio]{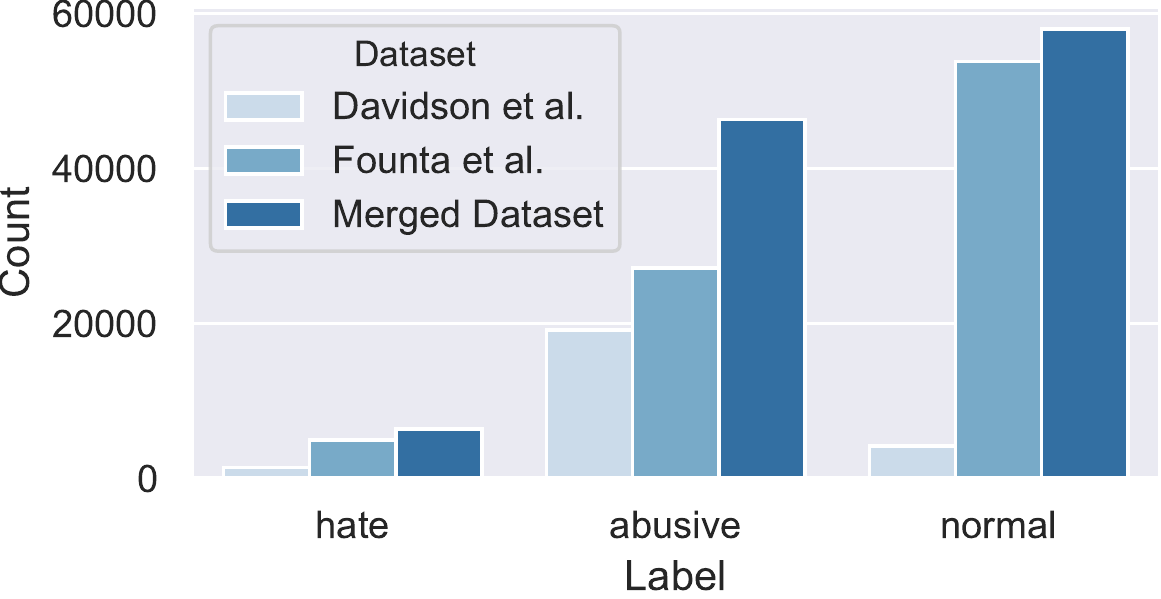}
        \caption{Counts for the respective classes in the datasets.}
        \label{fig:counts}
    \end{center}
\end{figure}

\begin{figure}[h]
    \begin{center}
    \includegraphics[width=\columnwidth,keepaspectratio]{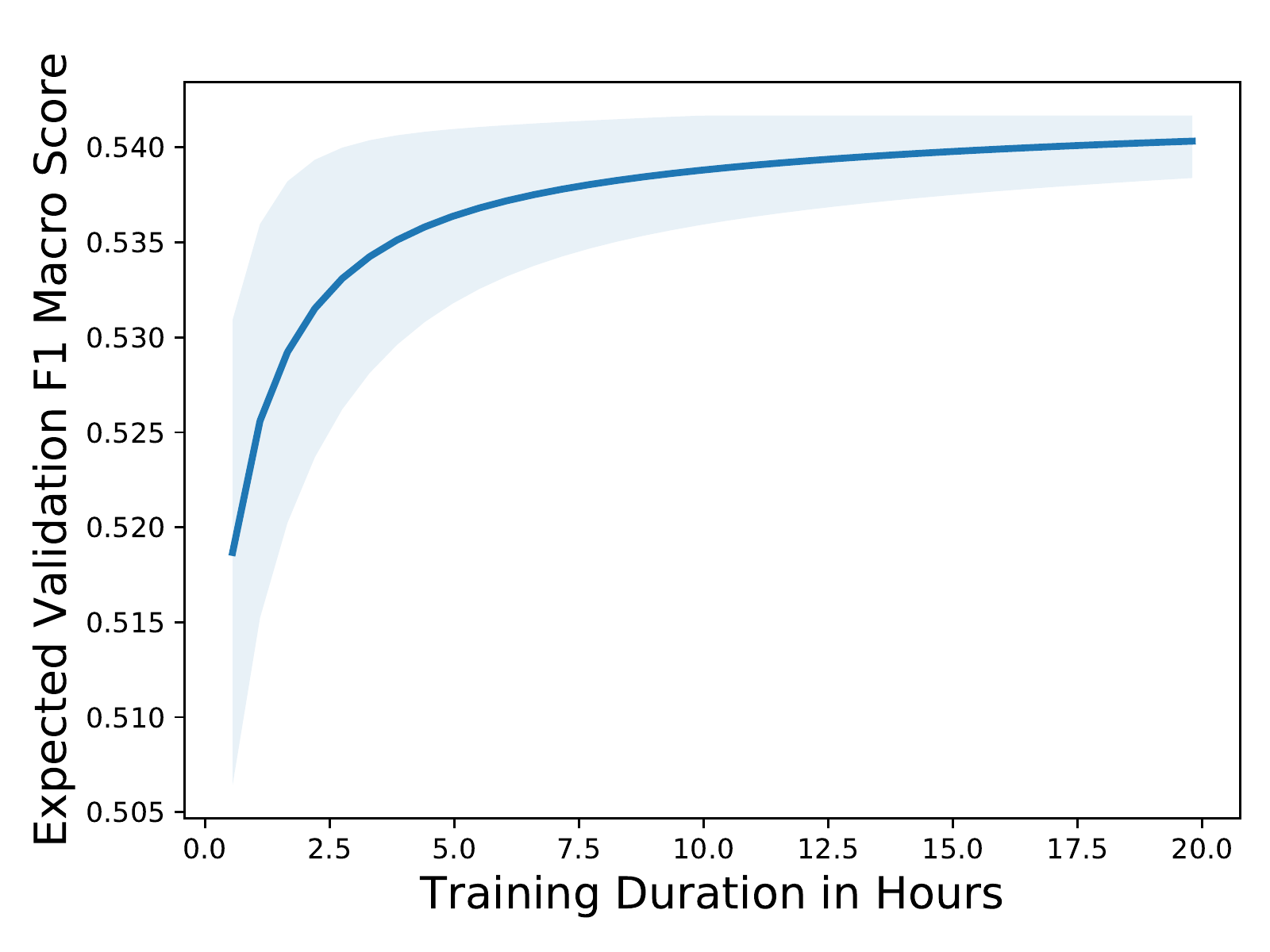}
    \caption{Expected validation performance of performed hyperparameter grid search.}
    \label{fig:Expected Validation Performance}
    \end{center}
\end{figure}

\begin{figure}[h]
    \begin{subfigure}{1\linewidth}
        \includegraphics[width=0.9\columnwidth]{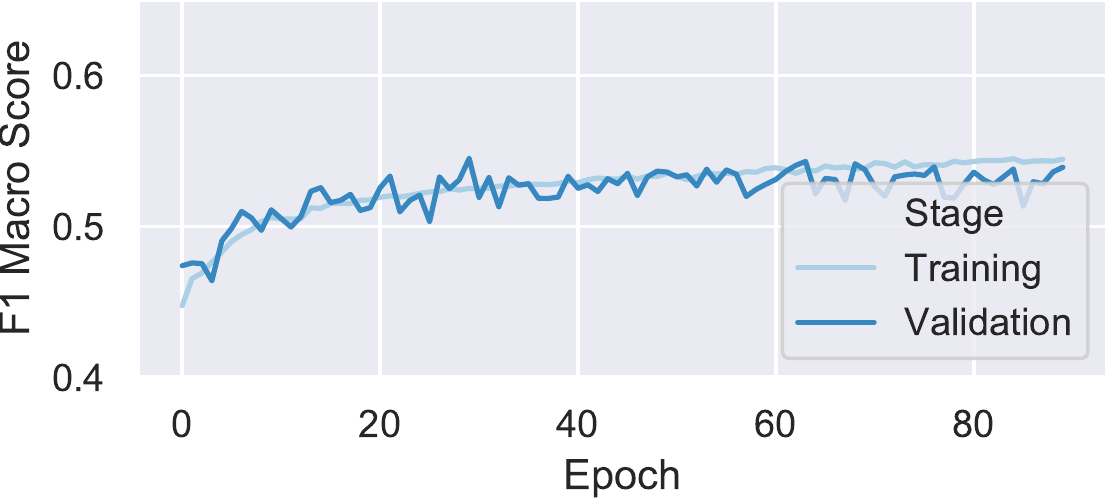} 
        \caption{1D Model with BERT preprocessing.}
        \label{fig:f1:1d-bert}
    \end{subfigure}
        \begin{subfigure}{1\linewidth}
        \includegraphics[width=0.9\columnwidth]{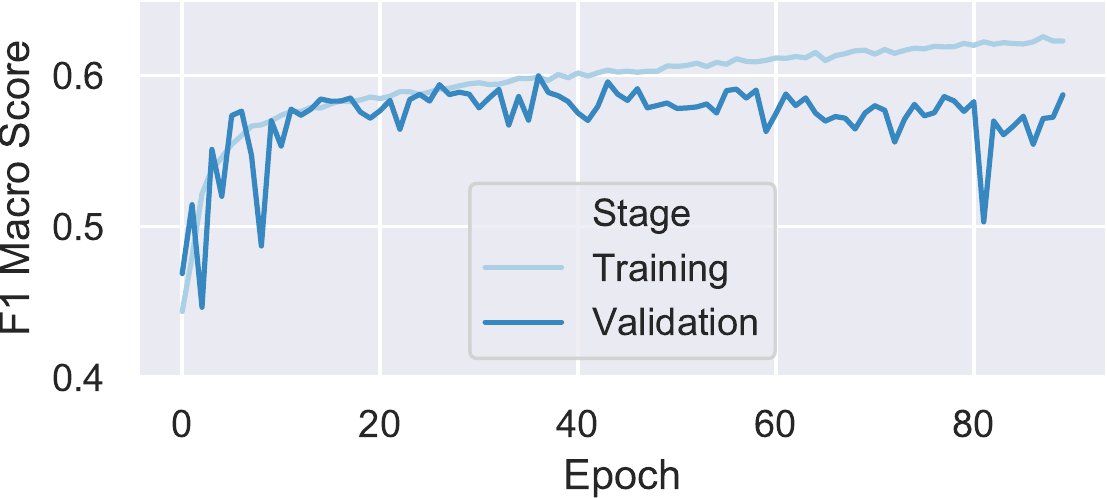}
        \caption{2D Model with additional dictNN preprocessing.}
        \label{fig:f1:2d-dictnn}
    \end{subfigure}
    \caption{F1 macro score plot of the 1D and 2D model over 90 epochs.}
    \label{fig:f1}
\end{figure}

\begin{figure}[h]
    \begin{subfigure}{1\linewidth}
        \includegraphics[width=0.9\columnwidth]{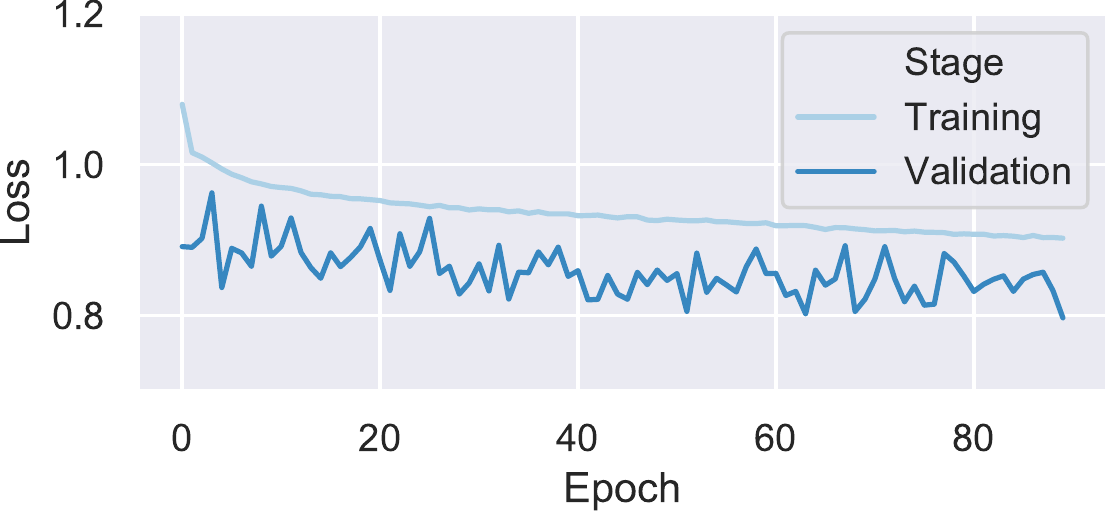} 
        \caption{1D Model with BERT preprocessing.}
        \label{fig:loss:1d}
    \end{subfigure}
        \begin{subfigure}{1\linewidth}
        \includegraphics[width=0.9\columnwidth]{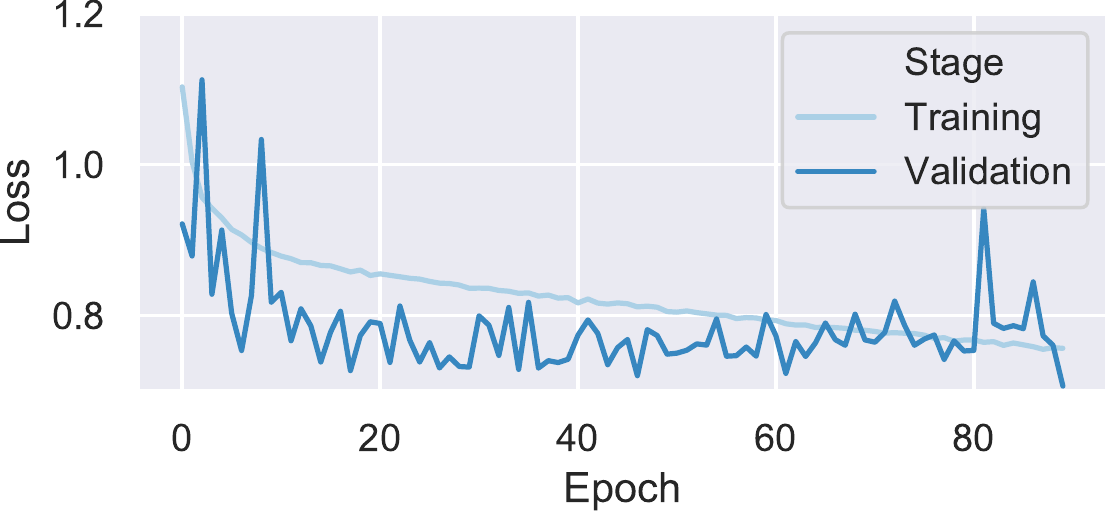}
        \caption{2D Model with additional dictNN preprocessing.}
        \label{fig:loss:2d}
    \end{subfigure}
    \caption{Loss plot of the 1D and 2D model over 90 epochs.}
    \label{fig:loss}
\end{figure}

\begin{figure}[h]
    \begin{subfigure}{1\linewidth}
        \includegraphics[width=0.9\columnwidth]{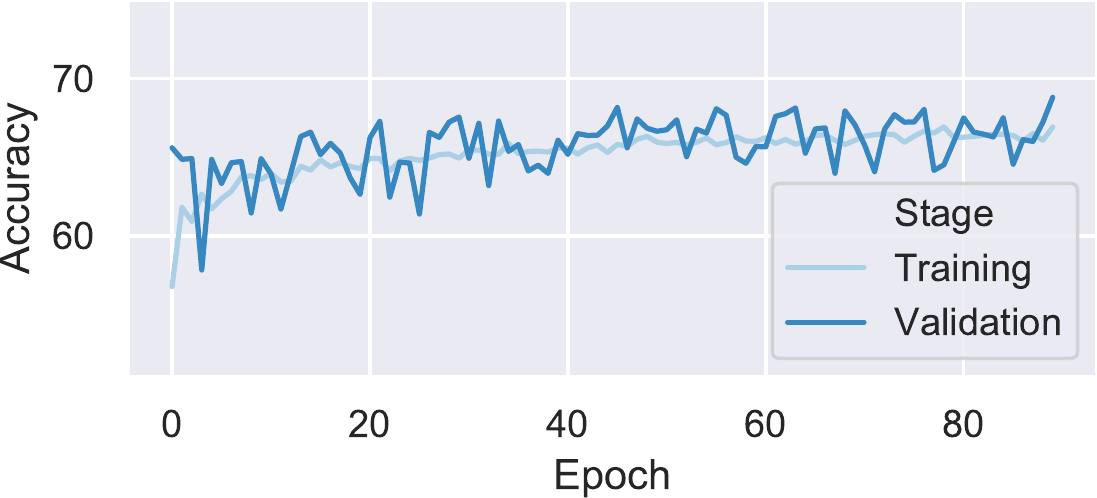} 
        \caption{1D Model with BERT preprocessing.}
        \label{fig:accuracy:1d}
    \end{subfigure}
        \begin{subfigure}{1\linewidth}
        \includegraphics[width=0.9\columnwidth]{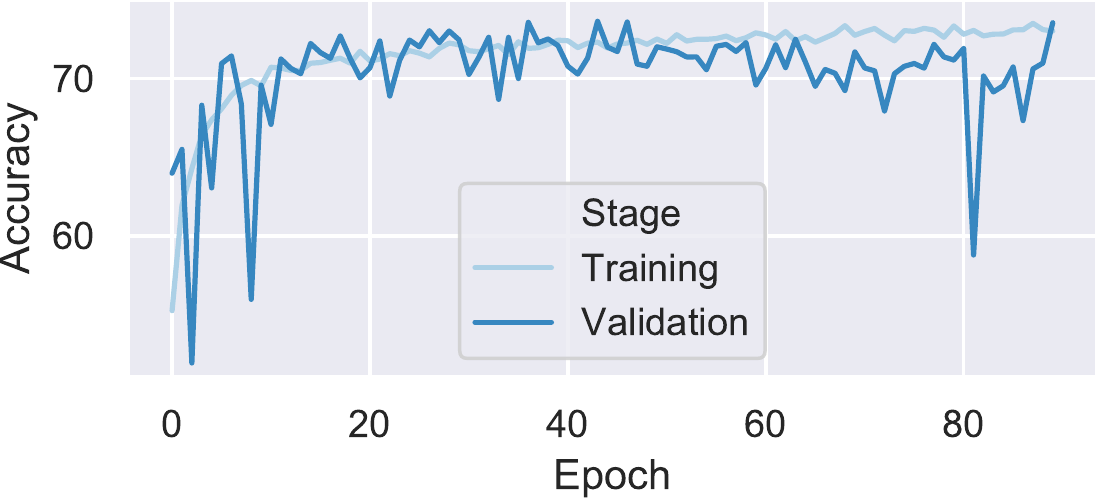}
        \caption{2D Model with additional dictNN preprocessing.}
        \label{fig:accuracy:2d}
    \end{subfigure}
    \caption{Accuracy plot of the 1D and 2D model over 90 epochs.}
    \label{fig:accuracy}
\end{figure}

\noindent
\begin{table}[h]
    \resizebox{\columnwidth}{!}{%
        \small\addtolength{\tabcolsep}{-3pt}
        \begin{tabular}{c r c c c c c}
        \toprule
        
        & \multicolumn{2}{c}{\bfseries Precision \hspace{0.05cm}} &
        \multicolumn{2}{c}{\bfseries Recall \hspace{0.05cm}} &
        \multicolumn{2}{c}{\bfseries F1 Score} \\
        \cmidrule(lr){2-3} \cmidrule(lr){4-5} \cmidrule(lr){6-7}
        & 1D & 2D \hspace{0.05cm} & 1D & 2D \hspace{0.05cm} & 1D & 2D \\
        \cmidrule(lr){1-7}
        \textit{hateful}    &  0.21 &  0.29 \hspace{0.05cm} &  0.26 &  0.28 \hspace{0.05cm} &  0.23 &  0.29 \\
        \textit{abusive}    &  0.70 &  0.76 \hspace{0.05cm} &  0.63 &  0.68 \hspace{0.05cm} &  0.66 &  0.72 \\
        \textit{normal}     &  0.72 &  0.76 \hspace{0.05cm} &  0.76 &  0.84 \hspace{0.05cm} &  0.74 &  0.80 \\
        \textbf{Accuracy}   &       &                       &       &                       &  0.68 &  0.73 \\
        \textbf{Macro Avg}  &  0.54 &  0.60 \hspace{0.05cm} &  0.55 &  0.60 \hspace{0.05cm} &  0.55 &  0.60 \\
        \textbf{Micro Avg}  &  0.68 &  0.73 \hspace{0.05cm} &  0.68 &  0.73 \hspace{0.05cm} &  0.68 &  0.73 \\
        \bottomrule
        
        \end{tabular}
    }
    \caption{Classification report of the best 1D model with BERT preprocessing (epoch 29) and 2D model with additional dictNN preprocessing (epoch 36) on the validation set.}
    \label{tab: classification report on validation}
\end{table}

\begin{figure}[h]
    \begin{subfigure}{1\linewidth}
        \includegraphics[width=0.9\columnwidth]{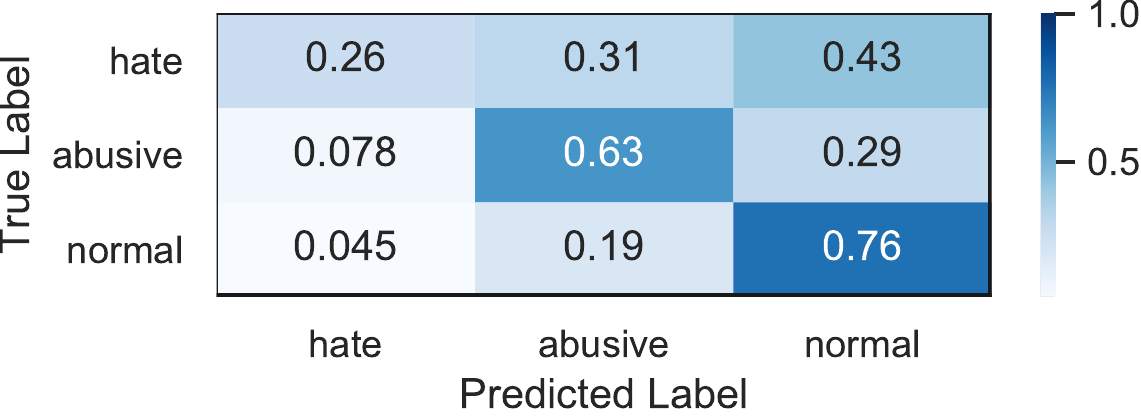} 
        \caption{1D Model with BERT preprocessing.}
        \label{fig:confusion-on-validation-set:1d}
    \end{subfigure}
        \begin{subfigure}{1\linewidth}
        \includegraphics[width=0.9\columnwidth]{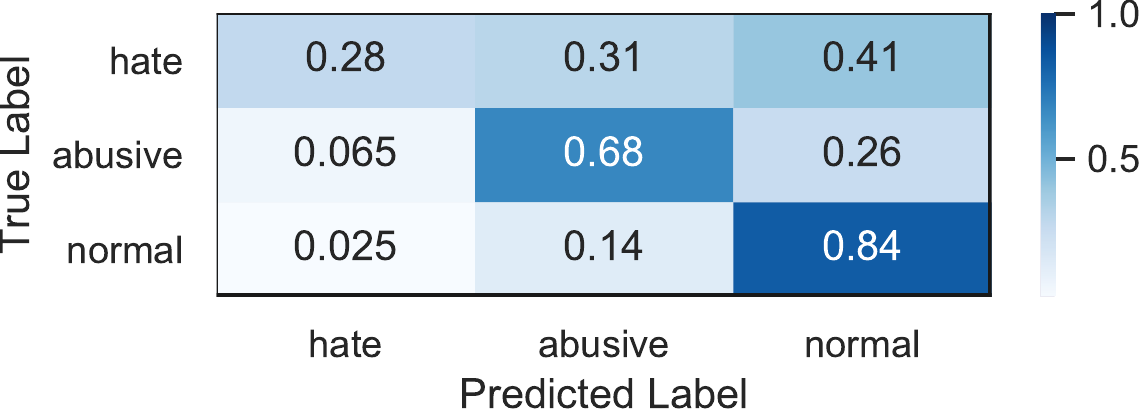}
        \caption{2D Model with additional dictNN preprocessing.}
        \label{fig:confusion-on-validation-set:2d}
    \end{subfigure}
    \caption{Confusion matrix of the 1D model (epoch 29) and the 2D model (epoch 36) on the validation set.}
    \label{fig:confusion-on-validation-set}
\end{figure}

\begin{table}[h]
    \begin{tabular}{p{\columnwidth}}
    \toprule
\multicolumn{1}{c}{\textbf{Example Tweet 1}}\label{Example Tweet 1}\\
"\textit{@[user1] @[user2] So the only way to not having children gassed to death is to topple Assad}"\\ 
\multicolumn{1}{c}{\textbf{Example Tweet 2}}\label{Example Tweet 2} \\ 
"\textit{I just watched a clip with a bunch of white folks shouting n[***]a \& going crazy to songs about black guys murdering one another \& it made me so sad.}"\\ 
\multicolumn{1}{c}{\textbf{Example Tweet 3}}\label{Example Tweet 3} \\
"\textit{I WARNED YOU CAUCASIAN, COUSIN F[**]KING, INCEST F[**]KS NOT TO VOTE RED NOW LOOK AT WHAT HAPPENED \#ThatsitMurica}"\\ \bottomrule
\caption{Synthetic tweets mirroring examples from the test set.}
\label{tab: Example tweets}
\end{tabular}
\end{table}

\end{document}